\documentclass[11pt,a4paper]{article}

\usepackage{times,latexsym}
\usepackage{url}
\usepackage[T1]{fontenc}
\usepackage[utf8]{inputenc}
\usepackage{amsmath,amssymb}
\usepackage{tabularx,booktabs,multirow}
\usepackage{xcolor,colortbl}
\usepackage{algorithm}
\usepackage{algpseudocode}
\usepackage[font=small,labelsep=colon]{caption}
\algrenewcommand{\algorithmicrequire}{\textbf{Input:}}
\algrenewcommand{\algorithmicensure}{\textbf{Output:}}
\definecolor{ourrow}{RGB}{235,243,250}
\definecolor{gain}{RGB}{0,128,96}
\usepackage{microtype}
\usepackage{inconsolata}
\usepackage{graphicx}
\usepackage{flafter}

\usepackage[acceptedWithA]{tacl2021v1}
\hypersetup{pdfauthor={Rui Ding, Yuren Mao, Yunjun Gao}}

\title{\textbf{MemTab}: An Efficient Memory Chunking and Indexing Method Based on Position-Space Combinatorial Optimization}
\author{%
  \textbf{Rui Ding}\textsuperscript{1},
  \textbf{Yuren Mao}\textsuperscript{1}\Thanks{Corresponding author.}\hspace{3pt},
  \textbf{Yunjun Gao}\textsuperscript{1}\\
  \textsuperscript{1}School of Software Technology, Zhejiang University\\
  \texttt{\{rui.ding, yuren.mao, gaoyj\}@zju.edu.cn}%
}
\date{}

\begin{document}
\maketitle

\begin{abstract}
Long-term memory lets autonomous agents sustain coherent interactions across extended horizons, yet most existing methods cut the stream at a fixed token threshold and then rewrite the chunks into summaries, graphs, or trees.
In practice, fixed cuts ignore where topics actually shift in unstructured memory text, and rewriting irreversibly discards original content.
To address both, we introduce \textbf{Memory Tabs}, which segments a long memory by semantics, pairs each chunk with a concise retrieval cue, and leaves the source text unmodified.
Then, we formulate its construction as a combinatorial optimization problem, and propose \textbf{MemTab}, a $154\text{M}$ dual-pointer network that chunks a memory and builds an index for each chunk without any LLM call.
On LoCoMo and LongMemEval-S, MemTab reaches $73.83\%$ and $84.20\%$ QA accuracy and $90.71\%$ and $96.39\%$ Recall@10, outperforming other memory baselines.
The MemTab model weights are available at \url{https://github.com/ZJU-LLMs/MemTab}.
\end{abstract}

\section{Introduction}

Long-term memory is critical for autonomous agents to sustain coherent interactions across extended horizons \citep{wang2024survey}. However, most existing memory management methods build such a store by first cutting the stream into pieces with a fixed token threshold and then operating on the resulting chunks \citep{packer2023memgpt,kang2025memoryos}. This recipe has two problems. First, a fixed-threshold partition does not fit unstructured raw memory text, where topical shifts carry no explicit structural marker that a threshold could align with. Second, many methods rewrite the chunks they obtain, consolidating them into summaries or rebuilding them into graphs and trees \citep{chhikara2025mem0,sarthi2024raptor,fang2026lightmem}. Such rewriting inevitably discards original information and detaches the memory from its source \citep{bao2026rawrag}. Hence, an effective memory structure needs boundaries that follow content itself while the source text remains intact.

\begin{figure}[t]
\centering
\includegraphics[width=0.8\columnwidth]{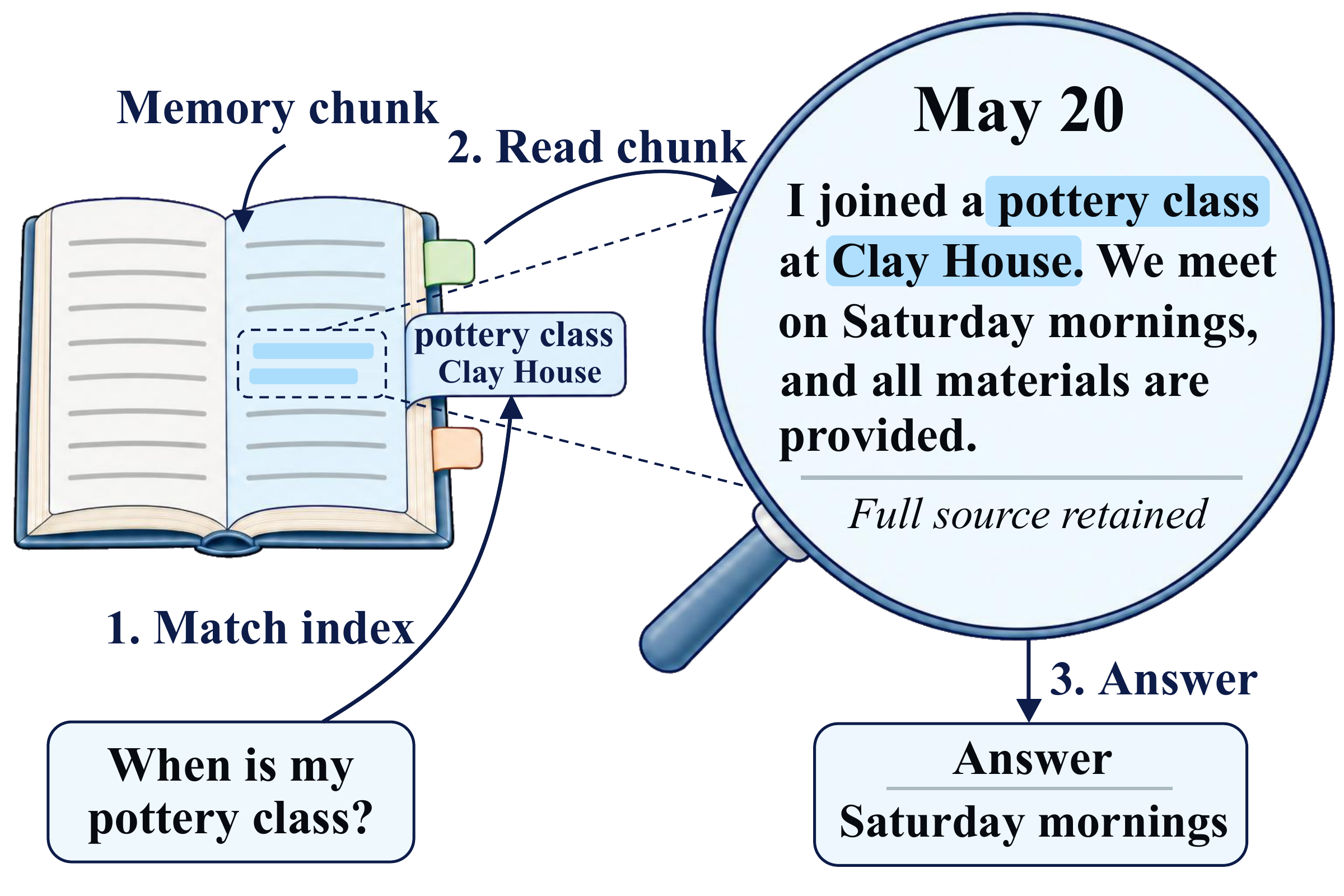}
\caption{\textbf{Memory Tabs as sticky book tabs.} Queries match extracted lexical cues to retrieve complete source chunks.}
\label{fig:memory_tabs}
\end{figure}

We therefore propose \textit{Memory Tabs}, a novel memory structure that meets both requirements. Like sticky tabs on the edge of a book, a memory tab splits a long memory into cohesive chunks and pairs each chunk with a concise retrieval cue (Figure~\ref{fig:memory_tabs}). This design is supported by the Hippocampal Memory Indexing Theory, in which cortical episodes are segmented and indexed by sparse hippocampal pointers for cue-driven recall \citep{teyler1986hippocampal}. Concretely, Memory Tabs segments raw text adaptively by semantics, attaches to each segment an index that locates it during retrieval, and reads back the original segment at recall time. The structure thus supplies well-formed retrieval units and precise retrieval cues while preserving the source text in full.

Despite these advantages, no existing method can construct Memory Tabs end-to-end and efficiently. Prevailing pipelines segment first and index afterwards, so the index construction cannot inform where boundaries fall while boundaries cannot adjust to keep retrieval cues distinct. Moreover, indexing relies on open-vocabulary autoregressive generation, whose search space is exponential in output length over the full vocabulary.

To fill this gap, we formulate end-to-end Memory Tab construction as \textbf{Anchored Sequence Partitioning (ASP)}, a combinatorial optimization problem, and propose a lightweight dual-pointer network based on Pointer Networks \citep{vinyals2015pointer}. The network sequentially selects segment boundaries and in-chunk lexical anchors, constructing Memory Tabs with minimal overhead.

To validate the effectiveness of MemTab, we conduct experiments on the LoCoMo \citep{maharana2024locomo} and LongMemEval-S \citep{wu2024longmemeval} benchmarks. MemTab attains the best overall QA accuracy on both benchmarks, and its advantage over raw-text window retrieval concentrates on multi-hop and multi-session questions that require assembling evidence across turns. It also improves source-grounded retrieval on both benchmarks, builds memories with no LLM tokens, and constructs them $12.7\times$--$506\times$ faster than LLM-indexed stores.

\begin{figure*}[t]
\centering
\includegraphics[width=\textwidth]{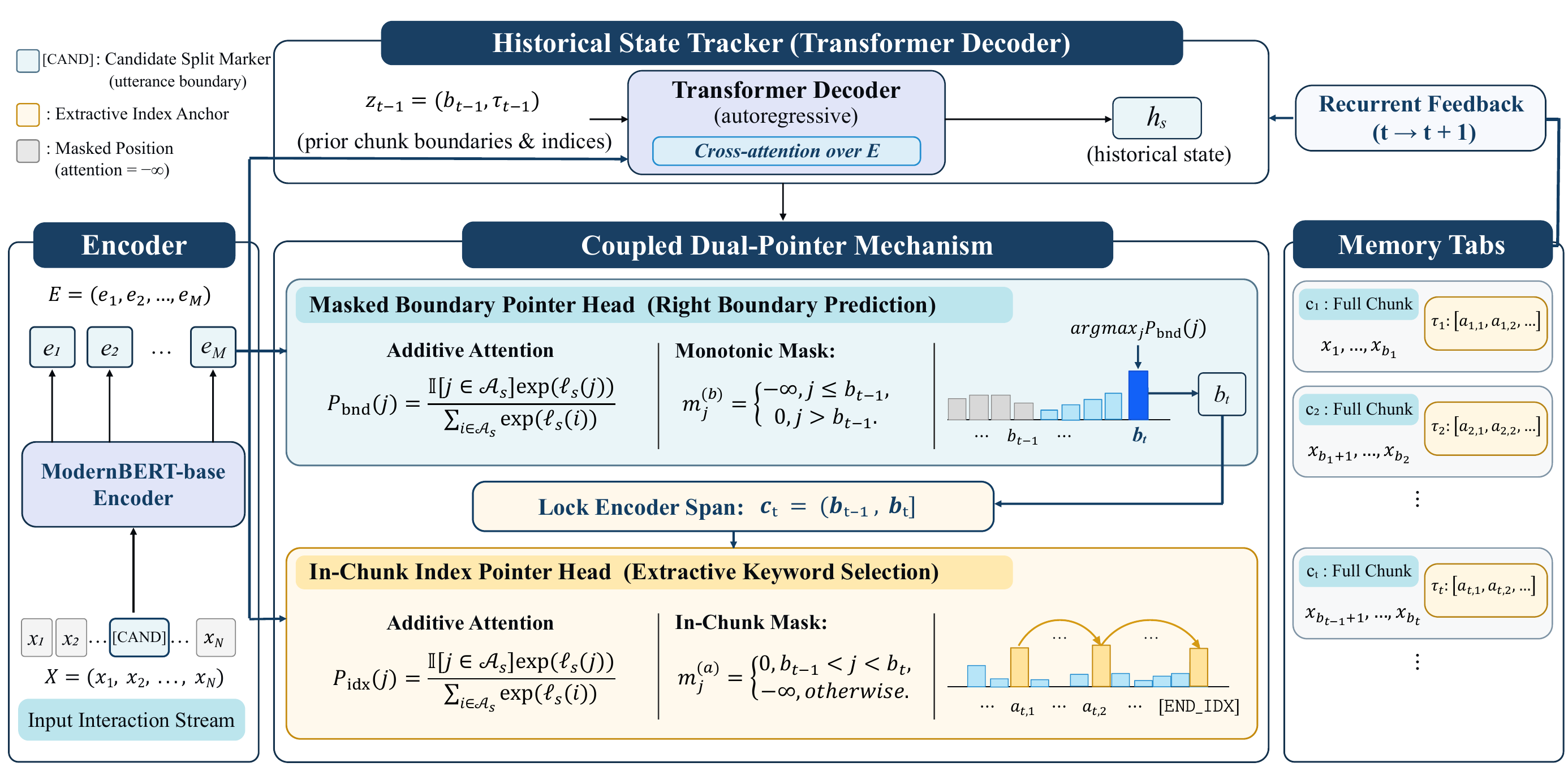}
\caption{\textbf{MemTab architecture.} A ModernBERT encoder contextualizes a source window annotated with candidate split markers; a two-layer Transformer decoder tracks the decisions made so far, and two pointer heads alternately select each chunk boundary and its in-chunk lexical anchors, producing the memory tabs pair $(c_t, \tau_t)$.}
\label{fig:method_architecture}
\end{figure*}

\section{Related Work}
\label{sec:related_work}

\paragraph{Long-Term Memory for LLM Agents.}
Sustaining multi-session coherence requires organizing interaction histories into structured external stores \citep{park2023generative,wang2024survey}. Pipeline systems such as MemGPT \citep{packer2023memgpt}, MemoryBank \citep{zhong2024memorybank}, MemoryOS \citep{kang2025memoryos}, and LightMem \citep{fang2026lightmem} manage episodic buffers through heuristic window thresholds, standalone topic segmentation, or scheduled reflection. Meanwhile, fact-centric and graph-centric frameworks including Mem0 \citep{chhikara2025mem0}, A-MEM \citep{xu2025amem}, HippoRAG \citep{gutierrez2024hipporag}, and CogMem \citep{liao2026cogmem} extract relational triples or memory notes from predefined dialogue turns. However, these pipelines decouple unit segmentation from index construction and fix boundaries prior to memory abstraction. In contrast, MemTab formulates boundary partitioning and cue extraction as a single \textbf{Anchored Sequence Partitioning} objective. Each resulting memory unit forms a cohesive discourse episode paired with discriminative retrieval anchors.

\paragraph{Context Segmentation and Memory Indexing.}
Long-context retrieval depends on both stream partitioning and segment indexing \citep{maharana2024locomo,wu2024longmemeval}. Traditional segmentation methods include lexical cohesion models \citep{hearst1997texttiling} and neural segmenters \citep{koshorek2018text,xing2021improving}. These models detect topical shifts solely from local semantic similarity and ignore downstream retrieval utility. Conversely, recent indexing methods enrich representations through abstractive rewrites, including tree summarization in RAPTOR \citep{sarthi2024raptor}, gist compression in ReadAgent \citep{lee2024readagent}, and proposition generation in Dense X Retrieval \citep{chen2024dense}. Yet autoregressive index generation incurs high latency and risks factual hallucinations \citep{maynez2020faithfulness}. MemTab bridges segmentation and indexing without generative rewriting. It extracts lexical anchors directly from source positions within dynamic boundaries, ensuring factual faithfulness and exact provenance.

\paragraph{Pointer Networks and Combinatorial Sequence Selection.}
Pointer Networks \citep{vinyals2015pointer} approximate intractable combinatorial problems by repurposing attention distributions as pointers over discrete input positions. In natural language processing, pointer mechanisms ground outputs in source indices for token copying \citep{gu2016incorporating,see2017get}, parsing and span extraction \citep{ma2018stack,yan2021unified}, extractive summarization \citep{jadhav2018extractive}, and text segmentation in SegBot \citep{li2018segbot}. However, existing pointer models operate on a single structural level. Segmentation models \citep{li2018segbot} predict split boundaries without retrieval cues, whereas extractive summarizers \citep{jadhav2018extractive} select salient elements from static documents without partitioning streams. MemTab resolves this limitation with a dual-pointer architecture. Our model interleaves monotonic boundary pointing with masked in-chunk anchor selection in a single pass.

\section{Problem Formulation: Anchored Sequence Partitioning}
\label{sec:problem_formulation}

Given a continuous interaction stream, our goal is to jointly partition the stream into cohesive discourse chunks and extract source-grounded retrieval cues. In this section, we first establish the information-theoretic motivation for joint optimization in the position space (\S\ref{sec:preliminaries}), and then formulate memory tab construction as an NP-hard Anchored Sequence Partitioning problem (\S\ref{sec:formulation}).

\subsection{Preliminaries and Theoretical Motivation}
\label{sec:preliminaries}

\paragraph{Notations.}
Let $X = (x_1, \dots, x_N)$ denote a continuous, unstructured interaction stream of length $N$. Each token $x_i \in \mathcal{V}$ is indexed by a discrete position in $\mathcal{P} = \{1, \dots, N\}$. We aim to transform $X$ into a set of memory tabs $\mathcal{M} = (C, I) = \{(c_t, \tau_t)\}_{t=1}^T$ with an adaptive number of memory units $T \ll N$. Here, $C = \{c_1, \dots, c_T\}$ denotes non-overlapping, exhaustive semantic chunks, and $I = \{\tau_1, \dots, \tau_T\}$ denotes the paired concise index tabs. Downstream memory access is exercised by queries $Q$ and measured by target responses $Y$.

\paragraph{Information-Theoretic Motivation.}
From an \textit{Information Bottleneck} perspective, an optimal memory structure $\mathcal{M} = (C, I)$ compresses the raw stream $X$ while maximizing the conditional mutual information $\mathcal{I}(Y; C, I \mid Q)$ for downstream answers $Y$ given queries $Q$. By the chain rule of mutual information, this objective decomposes into two coupled terms:
\begin{equation}
\label{eq:mi_decomp}
\mathcal{I}(Y; C, I | Q) = \mathcal{I}(Y; I | Q) + \mathcal{I}(Y; C | I, Q),
\end{equation}
The first term $\mathcal{I}(Y; I \mid Q)$ measures \textit{index recall utility}, while the second term $\mathcal{I}(Y; C \mid I, Q)$ measures \textit{chunk completeness utility}. However, existing decoupled pipelines fix index-agnostic chunks $\hat{C}$ before synthesizing indices $\hat{I}$. This staged design leaves a sub-optimality gap $\Delta_{\text{gap}} = \max_{C,I}\mathcal{I}(Y; C, I \mid Q) - \mathcal{I}(Y; \hat{C}, \hat{I} \mid Q) \geq 0$, which is positive whenever the fixed chunks trade one term against the other. Specifically, overly coarse chunks dilute index specificity and reduce $\mathcal{I}(Y; I \mid Q)$, whereas overly fine chunks sever cross-utterance context and degrade $\mathcal{I}(Y; C \mid I, Q)$. Meanwhile, jointly generating $(C, I)$ over an open vocabulary $\mathcal{V}$ requires searching an $\mathcal{O}(|\mathcal{V}|^L)$ output space. Crucially, we observe that both boundary segmentation and extractive indexing reduce to discrete position selections over $\mathcal{P} = \{1, \dots, N\}$. This observation transforms joint memory construction into a bounded \textbf{combinatorial optimization problem}.

\subsection{Combinatorial Formulation and Complexity}
\label{sec:formulation}

\paragraph{Problem Definition and Optimization Objective.}
In the position space $\mathcal{P}$, we parameterize $(C, I)$ as a trajectory of $T$ position tuples $Z = \{z_t = (b_t, a_t)\}_{t=1}^T$, where $b_t$ denotes the chunk's right boundary and $a_t = (a_{t,1}, \dots, a_{t,m_t})$ denotes its selected anchor positions. Any valid trajectory $Z \in \Omega(X)$ must satisfy two structural constraints: (i) \textit{Monotonic Boundary Progression}, where right boundaries $b_t$ strictly increase from $b_0 = 0$ to $b_T = N$ at sentence or utterance ends so that each span $(b_{t-1}, b_t]$ defines a contiguous chunk $c_t$; and (ii) \textit{In-Chunk Anchor Containment}, where anchors in $a_t$ are chosen in strictly increasing order within $(b_{t-1}, b_t]$ while preserving whole-word integrity, forming a tab $\tau_t$.

Over the feasible space $\Omega(X)$, we formulate memory tab construction as an \textbf{Anchored Sequence Partitioning (ASP)} problem that maximizes a global quality function $\mathcal{F}(Z \mid X)$:
\begin{equation}
\label{eq:comb_obj}
\begin{aligned}
\max_{Z \in \Omega(X)} &\, \mathcal{F}(Z \mid X) = \sum_{t=1}^{T} \phi_{\text{loc}}(b_{t-1}, b_t, a_t) \\[-3pt]
&\quad + \gamma \sum_{t=1}^{T} \sum_{k=1}^{t-1} \psi_{\text{glob}}(z_t, z_k),
\end{aligned}
\end{equation}
where $\gamma > 0$ balances local and global utilities. Here, the \textit{intra-chunk local utility} $\phi_{\text{loc}}(b_{t-1}, b_t, a_t)$ measures the topical cohesion of chunk $c_t$ and the representativeness of anchors $a_t$ in capturing local facts. Meanwhile, the \textit{inter-chunk global interaction} $\psi_{\text{glob}}(z_t, z_k)$ quantifies the semantic distinctiveness between current unit $z_t$ and historical units $z_k$ for $k < t$. This global term penalizes redundant or ambiguous retrieval cues across segments.

\paragraph{Computational Complexity and NP-Hardness.}
When $\gamma = 0$, Eq.~(\ref{eq:comb_obj}) reduces to an additive semi-Markov segmentation problem that can be solved in polynomial time via dynamic programming. However, when $\gamma > 0$, the interaction term $\psi_{\text{glob}}$ couples step $t$ with the entire decision history $z_{1:t-1}$. This coupling violates the Markov property and causes the search state space to grow exponentially.

\textbf{Theorem 1 (NP-Hardness).} \textit{For $\gamma > 0$, maximizing the ASP objective $\mathcal{F}(Z \mid X)$ over $\Omega(X)$ in Eq.~(\ref{eq:comb_obj}) is NP-hard.}

\textit{Proof Sketch.} We construct a polynomial-time reduction from the NP-hard \textbf{Maximum Independent Set (MIS)} problem \citep{karp1972reducibility} on an undirected graph $G = (V, E)$. By mapping each vertex to a fixed two-token chunk where anchor choices encode vertex selection ($\phi_{\text{loc}} \in \{0, 1\}$) and $\psi_{\text{glob}}$ assigns $-\infty$ penalty to adjacent selected vertices, any trajectory with $\mathcal{F}(Z \mid X) \ge 0$ corresponds one-to-one to a valid independent set of size $\mathcal{F}(Z \mid X)$. The complete proof is provided in Appendix~\ref{sec:appendix}.

\section{Dual-Pointer Network for Memory Tabs Construction}
\label{sec:method}

To tractably solve the NP-hard ASP objective in Eq.~(\ref{eq:comb_obj}), we propose \textbf{MemTab} (Figure~\ref{fig:method_architecture}), a $154\text{M}$-parameter dual-pointer network. Section~\ref{sec:architecture} presents the network architecture, and Section~\ref{sec:usage} describes how the pretrained model can be directly deployed for memory management.

\subsection{The MemTab Network}
\label{sec:architecture}

\paragraph{From Search to Pointer Decoding.}
Because exact optimization of ASP is NP-hard, we recast it as an autoregressive sequential decision process $P_\theta(Z \mid X)$ that builds the trajectory chunk by chunk. At chunk step $t$, the joint probability factorizes into boundary prediction followed by boundary-conditioned anchor extraction:
\begin{equation}
\label{eq:ar_factor}
\begin{aligned}
P_\theta(Z \mid X) = \prod_{t=1}^{T} &P_\theta(b_t \mid z_{1:t-1}, X) \\[-2pt]
&\cdot P_\theta(a_t \mid b_t, z_{1:t-1}, X).
\end{aligned}
\end{equation}
Since every decision thereby selects a position in $\mathcal{P}$, the decoded trajectory stays inside $\Omega(X)$ by construction; below, $b_t$ and $a_t$ denote the corresponding encoder positions, and stored source offsets recover their ASP coordinates. This sequential position-selection process is implemented Pointer Networks \citep{vinyals2015pointer}, which turn attention distributions into pointers over input positions.

\paragraph{Four-Step Execution.}
As shown in Figure~\ref{fig:method_architecture}, our network executes four sequential operations:
\begin{itemize}
    \item \textbf{Contextual Encoding:} We insert a dedicated \texttt{[INDEX\_SPLIT]} special token at each candidate sentence or utterance end of the source stream, obtaining $\widetilde X=(\widetilde x_1,\dots,\widetilde x_M)$. Feeding $\widetilde X$ into a bidirectional \textbf{ModernBERT-base} \citep{warner2024modernbert} encoder yields the contextualized states $E=(\mathbf{e}_1,\dots,\mathbf{e}_M)$.
    \item \textbf{Historical State Tracking:} A 2-layer Transformer decoder with hidden dimension $d=384$ and $6$ attention heads updates its state $\mathbf{h}_s$ at decoding step $s$. It applies causal self-attention over all prior boundary and index actions followed by cross-attention over $E$. Therefore, the resulting state $\mathbf{h}_s$ compactly summarizes the decision history $z_{1:t-1}$ and serves as a continuous neural proxy for $\psi_{\text{glob}}$.
    \item \textbf{Masked Boundary Pointing:} The \textit{Boundary Pointer} scores candidate split-marker positions with additive attention, $\ell_s(j)=\mathbf{v}_{\text{bnd}}^\top\tanh(\mathbf{W}_{\text{bnd}}\mathbf{h}_s+\mathbf{k}_j)$, where the key $\mathbf{k}_j$ fuses the projected encoder state $\mathbf{e}_j$ with a learnable action-type embedding (see Appendix~\ref{sec:app_masking} for the full scoring). Masks guarantee that only valid positions are selected, excluding illegal splits, e.g., positions at or before $b_{t-1}$. Greedy decoding then sets $b_t=\arg\max_j P_{\text{bnd}}(j)$.
    \item \textbf{In-Chunk Index Pointing:} The \textit{In-Chunk Index Pointer} applies the same additive scoring with its own parameters and selects source-token positions strictly between $b_{t-1}$ and $b_t$. It enforces increasing anchors and whole-word continuation while masking inserted markers and special tokens. Greedy decoding then selects $a_{t,r}=\arg\max_j P_{\text{idx}}(j)$ until $\texttt{END\_IDX}$ completes the tab.
\end{itemize}

\paragraph{Balanced Joint Training.}
We train the dual-pointer network end-to-end via supervised fine-tuning on teacher-annotated trajectories. Every training trajectory is verified for monotonic coverage, in-chunk containment, and whole-word alignment. Many index steps are deterministic, e.g., an unfinished word forces its continuation subword, so averaging the token-level loss over all steps would dilute the gradients on active lexical choices. We therefore normalize the index cross-entropy loss exclusively over non-deterministic decision steps $D$:
\begin{equation}
\label{eq:loss_v3}
\mathcal{L} = \mathcal{L}_{\text{bnd}} + \lambda \mathcal{L}_{\text{idx}}, \quad \mathcal{L}_{\text{idx}} = \frac{1}{|D|} \sum_{s \in D} \text{CE}_s,
\end{equation}
where $\mathcal{L}_{\text{bnd}}$ denotes the mean boundary cross-entropy loss and $\lambda = 2$ balances chunking and indexing supervision.

\paragraph{Rolling Inference.}
Interaction streams in lifelong agents often exceed the encoder's context window. We therefore use tail-preserving rolling inference, committing completed chunks while retaining the final visible source unit as lookahead in non-terminal windows. Specifically, the model may select $\texttt{OPEN}$ during boundary prediction to carry the uncommitted tail into the next window, where it is re-encoded with incoming context; once the final source unit has been committed, it may instead select $\texttt{EOS}$ to end the stream. Appendix~\ref{sec:app_rolling} gives the inference details.

\subsection{Memory Management with MemTab}
\label{sec:usage}

Once pretrained, MemTab can be deployed off-the-shelf for memory management without task-specific fine-tuning. Its two pointer heads not only work together to construct a complete Memory Tab store, but also operate independently to support standalone memory segmentation or indexing.

\paragraph{Joint Chunking and Indexing.}
When both pointer heads run together, MemTab processes continuously growing raw interaction streams into paired memory tabs $(c_t, \tau_t)$ in one pass without rebuilding prior memory. This is the default mode evaluated in our experiments (\S\ref{sec:exp_setup}). Compared with two-stage or LLM-rewritten memory pipelines, joint execution aligns chunk boundaries with retrieval cues to avoid the sub-optimality gap in Eq.~(\ref{eq:mi_decomp}), preserves lossless source text and eliminates LLM token overhead during memory construction.

\paragraph{Standalone Segmentation or Indexing.}
Either pointer head can also be invoked alone by masking the other phase during inference. Masking the index-pointing phase turns the Boundary Pointer into a semantic segmenter that partitions interaction streams into cohesive chunks $C$ for downstream fact extractors or graph builders. By replacing rigid window cuts that sever cross-turn context, boundary-only MemTab produces self-contained episodic units and provides cleaner inputs for downstream memory consolidation. Conversely, when chunk boundaries are fixed externally and the boundary-pointing phase is masked, the In-Chunk Index Pointer extracts lexical tabs $\tau_t$ for pre-segmented memory chunks. Unlike full-chunk embeddings where conversational filler dilutes key information, these extracted tabs provide sharper retrieval cues. Moreover, compared with generating abstractive indices via LLMs, index-only MemTab builds faithful indices at minimal computational overhead.

\begin{table*}[t]
\centering
\footnotesize
\setlength{\tabcolsep}{1.6pt}
\textbf{LoCoMo}\par\smallskip
\begin{tabularx}{\textwidth}{l *{6}{>{\centering\arraybackslash}X}}
\toprule
\multirow{2}{*}{\textbf{Method}} & \multicolumn{2}{c}{\textbf{Overall (1,540)}} & \textbf{Single-Hop} & \textbf{Multi-Hop} & \textbf{Temporal} & \textbf{Open-Domain} \\
\cmidrule(lr){2-3}
 & \textbf{Correct} & \textbf{Judge} & \textbf{(841)} & \textbf{(282)} & \textbf{(321)} & \textbf{(96)} \\
\midrule
NaiveRAG & 963 / 1,540 & 62.53 & 76.10 & 39.72 & 54.52 & 37.50 \\
RAW-RAG & \underline{1,127 / 1,540} & \underline{73.18} & \textbf{84.54} & \underline{51.06} & 70.72 & \underline{46.88} \\
Dense X & 867 / 1,540 & 56.30 & 71.22 & 36.88 & 38.63 & 41.67 \\
Mem0 & 732 / 1,540 & 47.53 & 56.00 & 28.01 & 45.48 & 37.50 \\
MemoryOS & 522 / 1,540 & 33.90 & 43.16 & 21.63 & 21.50 & 30.21 \\
LightMem & 930 / 1,540 & 60.39 & 65.99 & 35.46 & \underline{74.14} & 38.54 \\
CogMem & 879 / 1,540 & 57.08 & 66.83 & 29.08 & 60.75 & 41.67 \\
\midrule
\rowcolor{ourrow}
\textbf{MemTab (Ours)} & \textbf{1,137 / 1,540} & \textbf{73.83} & \underline{81.69} & \textbf{53.90} & \textbf{76.95} & \textbf{53.12} \\
\bottomrule
\end{tabularx}

\vspace{4pt}

\textbf{LongMemEval-S}\par\smallskip
\setlength{\tabcolsep}{3.0pt}
\begin{tabularx}{\textwidth}{l *{8}{>{\centering\arraybackslash}X}}
\toprule
\multirow{2}{*}{\textbf{Method}} & \multicolumn{2}{c}{\textbf{Overall (500)}} & \textbf{User} & \textbf{Assistant} & \textbf{Preference} & \textbf{Multi-sess.} & \textbf{Temporal} & \textbf{Knowledge} \\
\cmidrule(lr){2-3}
 & \textbf{Correct} & \textbf{Judge} & \textbf{(70)} & \textbf{(56)} & \textbf{(30)} & \textbf{(133)} & \textbf{(133)} & \textbf{(78)} \\
\midrule
NaiveRAG & 337 / 500 & 67.40 & \underline{92.86} & \underline{94.64} & 50.00 & 54.89 & 53.38 & 76.92 \\
RAW-RAG & \underline{406 / 500} & \underline{81.20} & \textbf{95.71} & \textbf{100.00} & 53.33 & 67.67 & \underline{78.20} & \textbf{93.59} \\
Dense X & 377 / 500 & 75.40 & \underline{92.86} & 87.50 & 36.67 & 65.41 & 69.17 & \textbf{93.59} \\
Mem0 & 281 / 500 & 56.20 & 90.00 & 5.36 & 6.67 & 63.91 & 56.39 & 67.95 \\
MemoryOS & 313 / 500 & 62.60 & 84.29 & 80.36 & 16.67 & 57.89 & 57.14 & 65.38 \\
LightMem & 374 / 500 & 74.80 & 91.43 & 16.07 & \textbf{70.00} & \underline{78.95} & \textbf{81.20} & 85.90 \\
CogMem & 306 / 500 & 61.20 & 90.00 & 76.79 & 16.67 & 54.89 & 58.65 & 56.41 \\
\midrule
\rowcolor{ourrow}
\textbf{MemTab (Ours)} & \textbf{421 / 500} & \textbf{84.20} & 91.43 & \textbf{100.00} & \underline{56.67} & \textbf{79.70} & \textbf{81.20} & \underline{89.74} \\
\bottomrule
\end{tabularx}
\caption{\textbf{End-to-end QA Judge accuracy (\%) on LoCoMo and LongMemEval-S across all reasoning categories.} Numbers in header parentheses denote query counts. \textbf{Bold} and \underline{underline} indicate the best and second-best scores under the unified Qwen3.8-27B reader and judge.}
\label{tab:main_qa}
\end{table*}

\begin{table*}[t]
\centering
\footnotesize
\setlength{\tabcolsep}{1.8pt}
\begin{tabularx}{\textwidth}{l >{\hsize=0.75\hsize\linewidth=\hsize\centering\arraybackslash}X >{\hsize=0.8\hsize\linewidth=\hsize\centering\arraybackslash}X >{\hsize=1.15\hsize\linewidth=\hsize\centering\arraybackslash}X >{\hsize=1.2\hsize\linewidth=\hsize\centering\arraybackslash}X >{\hsize=1.1\hsize\linewidth=\hsize\centering\arraybackslash}X >{\hsize=0.75\hsize\linewidth=\hsize\centering\arraybackslash}X >{\hsize=0.8\hsize\linewidth=\hsize\centering\arraybackslash}X >{\hsize=1.15\hsize\linewidth=\hsize\centering\arraybackslash}X >{\hsize=1.2\hsize\linewidth=\hsize\centering\arraybackslash}X >{\hsize=1.1\hsize\linewidth=\hsize\centering\arraybackslash}X}
\toprule
\multirow{2}{*}{\textbf{Method}} & \multicolumn{5}{c}{\textbf{LoCoMo}} & \multicolumn{5}{c}{\textbf{LongMemEval-S}} \\
\cmidrule(lr){2-6} \cmidrule(lr){7-11}
 & \textbf{R@5} & \textbf{R@10} & \textbf{MRR@10} & \textbf{nDCG@10} & \textbf{All-Evid.} & \textbf{R@5} & \textbf{R@10} & \textbf{MRR@10} & \textbf{nDCG@10} & \textbf{All-Evid.} \\
\midrule
NaiveRAG & 66.32 & 76.05 & 55.87 & 58.29 & 69.09 & 49.41 & 67.59 & 37.36 & 40.53 & 51.91 \\
RAW-RAG & \textbf{87.78} & \textbf{92.39} & \underline{81.84} & \underline{82.21} & \textbf{86.84} & \underline{79.24} & \underline{90.71} & \underline{60.83} & \underline{63.25} & \underline{82.77} \\
Dense X & 60.88 & 69.54 & 51.04 & 53.12 & 62.87 & 66.11 & 76.85 & 55.46 & 57.43 & 64.47 \\
MemoryOS & 27.90 & 31.25 & 26.71 & 27.75 & 26.65 & 60.66 & 76.32 & 58.71 & 57.87 & 64.47 \\
\midrule
\rowcolor{ourrow}
\textbf{MemTab (Ours)} & \underline{85.95} & \underline{90.71} & \textbf{83.07} & \textbf{82.93} & \underline{85.99} & \textbf{90.26} & \textbf{96.39} & \textbf{85.87} & \textbf{86.60} & \textbf{92.34} \\
\bottomrule
\end{tabularx}
\caption{\textbf{Source-grounded retrieval performance in percentage on LoCoMo and LongMemEval-S.} \textbf{All-Evid.} denotes the percentage of queries with all ground-truth evidence turns recovered within the top-10 chunks.}
\label{tab:main_retrieval}
\end{table*}

\section{Experiments}
\label{sec:experiments}

We evaluate MemTab across two long-term conversational memory benchmarks, detailing our experimental setup (\S\ref{sec:exp_setup}), end-to-end QA performance (\S\ref{sec:main_qa}), source-grounded retrieval quality (\S\ref{sec:retrieval}), memory construction efficiency (\S\ref{sec:efficiency}), and qualitative case analysis (\S\ref{sec:case_study}).

\subsection{Experimental Setup}
\label{sec:exp_setup}

\paragraph{Benchmarks and Metrics.}
We evaluate MemTab on two long-term conversational memory benchmarks. \textbf{LoCoMo} \citep{maharana2024locomo} contains $10$ multi-session dialogues with $1,540$ QA pairs and $1,527$ evidence-annotated queries. \textbf{LongMemEval-S} \citep{wu2024longmemeval} comprises $500$ isolated interaction histories across six reasoning categories with $470$ answerable queries. For end-to-end QA, we report \textbf{Judge Accuracy} on both benchmarks. For source-grounded retrieval, we measure \textbf{Recall@5}, \textbf{Recall@10}, \textbf{MRR@10}, \textbf{nDCG@10}, and \textbf{All-Evidence Recall}. For construction efficiency, we track \textbf{LLM Token Consumption} in thousands per store and \textbf{Wall-Clock Latency} in seconds per store.

\paragraph{Baselines.}
We compare MemTab against seven baselines. \textbf{NaiveRAG} tokenizes the dialogue into $256$-token chunks and retrieves the top-$10$ by embedding similarity. \textbf{RAW-RAG} \citep{bao2026rawrag} tokenizes the dialogue into $256$-token windows and retrieves the top-$128$ by embedding similarity, then reranks them to the top-$10$. \textbf{Dense X Retrieval} \citep{chen2024dense} generates atomic propositions via Flan-T5-large. We also evaluate four structured agent memory systems: \textbf{Mem0} \citep{chhikara2025mem0}, \textbf{MemoryOS} \citep{kang2025memoryos}, \textbf{LightMem} \citep{fang2026lightmem}, and \textbf{CogMem} \citep{liao2026cogmem}.

\paragraph{Implementation Details.}
Our $154\text{M}$-parameter dual-pointer network pairs a \texttt{ModernBERT-base} \citep{warner2024modernbert} encoder supporting an $8,192$-token window with a 2-layer Transformer decoder of hidden dimension $d=384$ and $6$ attention heads. To prevent test-set leakage, we train the model exclusively on out-of-domain meeting transcripts from \textbf{QMSum} \citep{zhong2021qmsum} with $585$ training and $137$ validation windows. We use \textbf{Qwen3.8-27B} at temperature $0$ for LLM operations and judging. We use \textbf{Qwen3-Embedding-4B} for dense retrieval and \textbf{Qwen3-Reranker-4B} to rerank \citep{zhang2025qwen3embedding}. For MemTab, retrieval first recalls the top-$100$ tabs $\tau_t$ by embedding similarity, and then the LLM reranks these candidates to select the most relevant indices, whose source chunks $c_t$ are read back for answer generation. Appendix~\ref{sec:app_eval_prompts} gives the prompt templates.

\begin{table}[t]
\centering
\footnotesize
\setlength{\tabcolsep}{2pt}
\begin{tabularx}{\columnwidth}{l *{4}{>{\raggedleft\arraybackslash}X}}
\toprule
\multirow{3}{*}{\textbf{Method}} & \multicolumn{2}{c}{\textbf{LoCoMo}} & \multicolumn{2}{c}{\textbf{LongMemEval-S}} \\
\cmidrule(lr){2-3} \cmidrule(lr){4-5}
 & \textbf{Tok.} & \textbf{Time} & \textbf{Tok.} & \textbf{Time} \\
 & (k/store) & (s/store) & (k/store) & (s/store) \\
\midrule
\multicolumn{5}{l}{\textit{Unindexed Fixed-Window Slicing}} \\
NaiveRAG & \textbf{0.00} & \textbf{2.8} & \textbf{0.00} & \textbf{11.0} \\
RAW-RAG & \textbf{0.00} & \underline{3.3} & \textbf{0.00} & \underline{13.7} \\
\midrule
\multicolumn{5}{l}{\textit{Structured \& Indexed Memory Construction}} \\
Dense X$^\dagger$ & \textbf{0.00} & 60.1 & \textbf{0.00} & 293.4 \\
Mem0 & 850.25 & 4957.4 & 967.59 & 17896.5 \\
MemoryOS & 199.94 & 1629.0 & 893.68 & 15528.6 \\
LightMem & 188.05 & 2989.4 & 51.29 & 1192.5 \\
CogMem & 262.88 & 829.2 & 193.47 & 2385.9 \\
\midrule
\rowcolor{ourrow}
\textbf{MemTab} & \textbf{0.00} & 9.8 & \textbf{0.00} & 94.0 \\
\bottomrule
\end{tabularx}
\caption{\textbf{Memory construction cost.} \textbf{Tok.} and \textbf{Time} denote construction LLM tokens in thousands per store and latency in seconds per store; lower is better for both. $^\dagger$Token counts exclude local Flan-T5 generation. Timing protocols are detailed in Appendix~\ref{sec:app_baseline_protocol}.}
\label{tab:construction_cost_v2}
\end{table}

\begin{figure*}[t]
\centering
\includegraphics[width=0.88\textwidth]{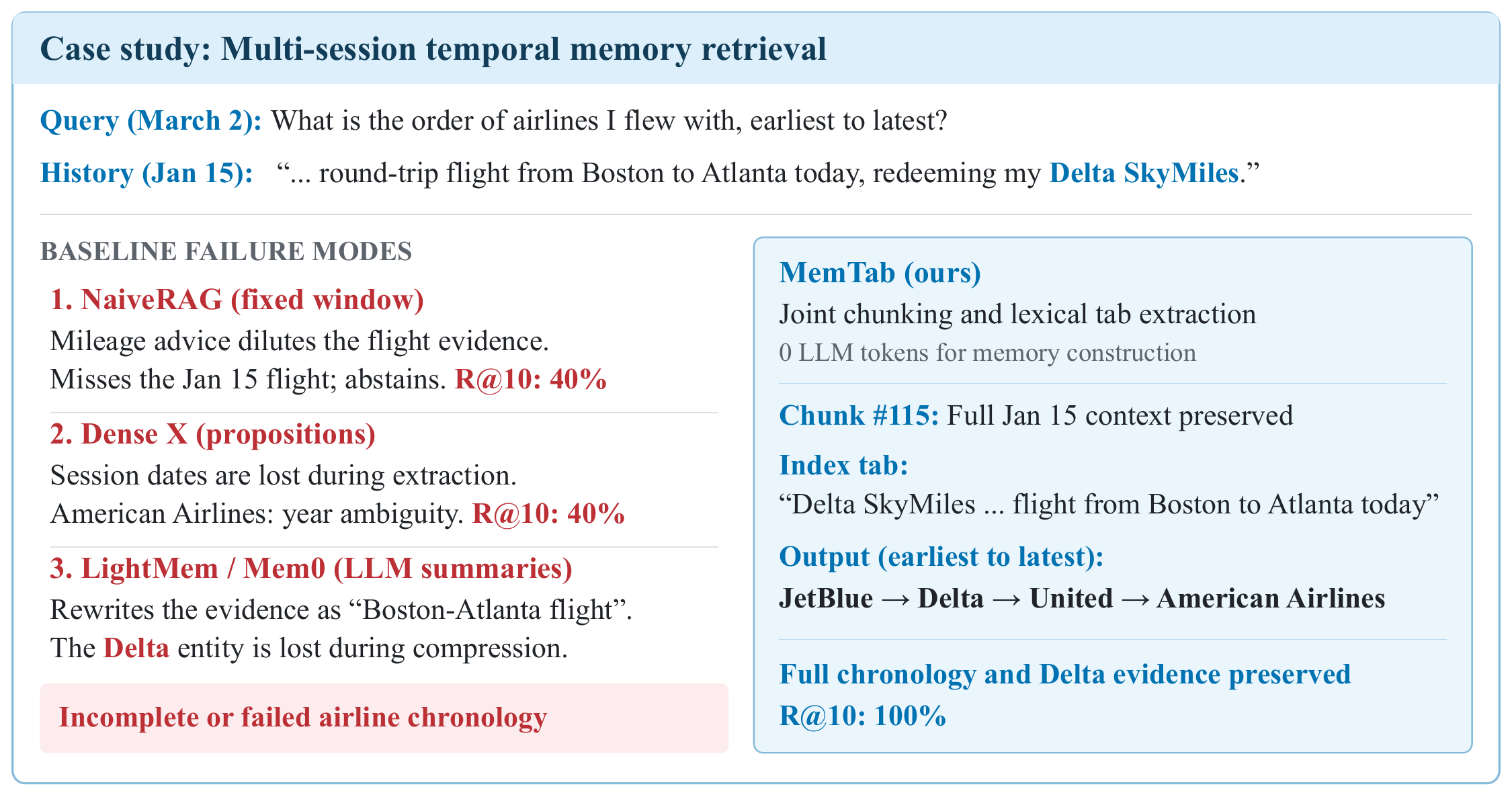}
\caption{\textbf{Multi-session temporal retrieval on a LongMemEval-S case.} The query asks for an ordering of four completed flights. Fixed-window retrieval dilutes the relevant turn, and abstractive memories drop the \textit{Delta} entity or its session date; MemTab keeps the Jan~15 context in the chunk and links it through a lexical tab, recovering the correct order.}\label{fig:case_study}
\end{figure*}

\subsection{Main Results on End-to-End QA}
\label{sec:main_qa}

\paragraph{Overall and Multi-Hop QA Performance.}
Table~\ref{tab:main_qa} shows that MemTab attains the highest overall QA Judge accuracy on both benchmarks, reaching $73.83\%$ on LoCoMo and $84.20\%$ on LongMemEval-S after zero-shot transfer from out-of-domain meeting transcripts. Although RAW-RAG is the strongest baseline, the two methods exhibit distinct strengths across reasoning categories. RAW-RAG performs well on single-hop and single-fact knowledge queries because a fixed $256$-token window rarely splits an answer contained within a single turn. In contrast, MemTab consistently ranks first on complex categories that require assembling evidence across multiple turns or sessions, including \textit{Multi-Hop}, \textit{Temporal}, and \textit{Open-Domain} on LoCoMo as well as \textit{Multi-session} and \textit{Temporal} on LongMemEval-S. By jointly aligning semantic boundaries with retrieval anchors, MemTab preserves cross-utterance coherence that fixed-window cuts sever. Meanwhile, LightMem leads on \textit{Preference} queries because its profile-style abstraction directly summarizes user traits, whereas Dense X struggles on temporal reasoning because isolated propositions strip conversational turn order and implicit session timestamps.

\paragraph{Robustness Against Catastrophic Forgetting.}
Comparing MemTab with two-stage abstractive memories exposes an inherent tension between targeted compression and context completeness, directly validating Eq.~(\ref{eq:mi_decomp}). LLM-consolidated systems such as LightMem excel on \textit{Temporal} and \textit{Preference} queries by prompting the 27B LLM to synthesize user-centric timelines and profiles. However, this schema-driven abstraction optimizes index utility $\mathcal{I}(Y; I \mid Q)$ only for anticipated user traits while irreversibly discarding surrounding context $\mathcal{I}(Y; C \mid I, Q)$. Consequently, on LongMemEval-S \textit{Assistant} queries that require recalling details previously stated by the AI, Mem0 and LightMem lose these details and fall to $5.36\%$ and $16.07\%$ accuracy. Because MemTab uses extractive tabs $\tau_t$ strictly as pointers to read back lossless source chunks $c_t$, it decouples retrieval compression from storage fidelity, reaching $100.00\%$ on \textit{Assistant} queries, tied with RAW-RAG and above NaiveRAG's $94.64\%$, while remaining strong on user-centric reasoning.

\subsection{Source-Grounded Retrieval Performance}
\label{sec:retrieval}

\paragraph{Provenance-Preserving Retrieval.}
To isolate retrieval quality from reader generation, Table~\ref{tab:main_retrieval} evaluates turn-level evidence recovery. The abstractive systems Mem0, LightMem, and CogMem are excluded because their consolidation pipelines rewrite multi-turn histories into synthetic notes detached from source turn IDs, precluding verifiable attribution. In contrast, restricting both boundaries and anchors to discrete input positions $\mathcal{P}$ guarantees $100\%$ source provenance for MemTab.

\paragraph{Retrieval Recall and Multi-Evidence Coverage.}
As shown in Table~\ref{tab:main_retrieval}, MemTab reaches $90.71\%$ and $96.39\%$ Recall@10 on LoCoMo and LongMemEval-S, outperforming all comparable baselines across every retrieval metric on LongMemEval-S. On LoCoMo, RAW-RAG attains slightly higher recall via redundant overlapping windows, yet MemTab achieves stronger top-rank precision on MRR@10 and nDCG@10. Both methods substantially surpass NaiveRAG, where conversational filler dilutes full-chunk embeddings, and Dense X, where isolated propositions lose coreference and speaker context. By anchoring embeddings on salient in-chunk tokens while returning complete episodes, MemTab sharpens top-rank discrimination and multi-turn evidence coverage, directly driving the multi-hop QA gains.

\subsection{Memory Construction Efficiency}
\label{sec:efficiency}

\paragraph{Zero-Token Overhead and Construction Speedup.}
Table~\ref{tab:construction_cost_v2} highlights the efficiency gain of reformulating memory construction from open-vocabulary generation to combinatorial position selection. Structured agent memories rely on iterative 27B LLM prompting over hundreds of thousands of tokens per store, while Dense X still incurs autoregressive decoding overhead from its $780\text{M}$ Flan-T5 model. In contrast, our $154\text{M}$ dual-pointer network predicts sparse boundary and anchor pointers in a single pass, consuming \textbf{no LLM tokens} and building a store in $9.8\text{s}$ on LoCoMo and $94.0\text{s}$ on LongMemEval-S---$12.7\times$ to $506\times$ faster than LLM-indexed stores while achieving the highest end-to-end QA accuracy.

\subsection{Qualitative Case Study}
\label{sec:case_study}

\paragraph{Qualitative Analysis.}
Figure~\ref{fig:case_study} illustrates a multi-session temporal query from LongMemEval-S that requires chronologically ordering four completed flights across $198$ chunks while excluding unflown plans. The observed failure modes vary across baseline types and are consistent with Eq.~(\ref{eq:mi_decomp}). Fixed-window retrieval in \textbf{NaiveRAG} misses the Jan~15 \textit{Delta} turn amid verbose mileage advice. Proposition-based retrieval in \textbf{Dense X} loses session dates during extraction, leaving the year of the \textit{American Airlines} flight ambiguous. LLM-based consolidation in \textbf{LightMem} and \textbf{Mem0} drops the \textit{Delta SkyMiles} reference when rewriting the Jan~15 turn. MemTab groups each flight mention with its local discourse context while extracting discriminative lexical anchors such as \textit{Delta SkyMiles}, recovering all four flights in the exact chronological order.

\section{Conclusion}
\label{sec:conclusion}

We introduced \textbf{Memory Tabs} and formulated their end-to-end construction as an NP-hard \textbf{Anchored Sequence Partitioning} problem over token positions. To solve it efficiently, we proposed \textbf{MemTab}, a $154\text{M}$-parameter dual-pointer network that jointly segments interaction streams into cohesive chunks and extracts in-chunk lexical tabs without generative rewriting. Across LoCoMo and LongMemEval-S, MemTab outperforms fixed-window and structured memory baselines on both end-to-end QA and source-grounded retrieval while constructing memory stores with zero LLM tokens and orders-of-magnitude lower latency.

\bibliographystyle{acl_natbib}
\bibliography{custom}

\appendix

\section{Formal Proof of Theorem 1}
\label{sec:appendix}

We present the complete formal proof of \textbf{Theorem~1}, which establishes the NP-hardness of maximizing the Anchored Sequence Partitioning objective $\mathcal{F}(Z \mid X)$ in Eq.~(\ref{eq:comb_obj}) for any global interaction weight $\gamma > 0$.

\paragraph{Theorem 1.} \textit{For $\gamma > 0$, maximizing the objective $\mathcal{F}(Z \mid X)$ over the feasible space $\Omega(X)$ subject to monotonic boundary progression and in-chunk anchor containment is NP-hard.}

\paragraph{Proof.} We construct a polynomial-time reduction from the decision version of the NP-hard Maximum Independent Set problem \citep{karp1972reducibility}. An instance of Maximum Independent Set is defined by an undirected graph $G = (V, E)$ with $T$ vertices $V = \{v_1, v_2, \dots, v_T\}$, an edge set $E \subseteq V \times V$, and an integer threshold $K \le T$. The decision problem asks whether there exists a vertex subset $S \subseteq V$ of cardinality $|S| \ge K$ such that no two vertices $u, v \in S$ share an edge in $E$.

\noindent \textbf{Reduction Construction.} Given $G = (V, E)$, we construct an equivalent instance $(X, \Omega(X), \phi_{\text{loc}}, \psi_{\text{glob}})$ in $\mathcal{O}(T^2)$ time through four steps:
\begin{enumerate}
    \item \textbf{Sequence Construction:} We construct a token sequence $X$ of length $N = 2T$ composed of $T$ contiguous two-token gadgets $I_t = \{2t-1, 2t\}$ for $t \in \{1, \dots, T\}$, where each gadget corresponds to vertex $v_t \in V$. Selecting the odd position $2t-1$ encodes including vertex $v_t$ in the independent set, whereas selecting the even position $2t$ encodes excluding $v_t$. We place candidate split boundaries at all even positions so that $\mathcal{P}_{\text{cand}} = \{2, 4, \dots, 2T\}$.
    \item \textbf{Feasible Space and Boundary Enforcement:} To lock chunk boundaries at the end of each gadget, we assign a utility of $-\infty$ to any trajectory where $b_t \neq 2t$ for any $t \in \{1, \dots, T\}$. Consequently, any valid trajectory achieving $\mathcal{F}(Z \mid X) > -\infty$ must partition $X$ into exactly $T$ two-token chunks $c_t = X[2t-1 : 2t]$ with right boundaries $b_t^* = 2t$.
    \item \textbf{Local Utility $\phi_{\text{loc}}$:} Within each locked chunk $c_t = X[2t-1 : 2t]$, we define the local utility of anchor selection $a_t$ as:
    \begin{equation}
    \phi_{\text{loc}}(2t{-}2, 2t, a_t) =
    \begin{cases}
    1, & a_t = \{2t{-}1\}, \\
    0, & a_t = \{2t\}, \\
    -\infty, & \text{otherwise}.
    \end{cases}
    \end{equation}
    \item \textbf{Global Interaction $\psi_{\text{glob}}$:} For any chunk pair $k < t$, if $(v_k, v_t) \in E$ and both active anchors $a_k = \{2k-1\}$ and $a_t = \{2t-1\}$ are selected, we impose an infinite collision penalty:
    \begin{equation}
    \psi_{\text{glob}}(z_t, z_k) =
    \begin{cases}
    -\infty, & \!\begin{aligned}[t]
               &(v_k, v_t) \in E, \\[-2pt]
               &a_k=\{2k{-}1\}, \\[-2pt]
               &a_t=\{2t{-}1\},
               \end{aligned} \\
    0, & \text{otherwise}.
    \end{cases}
    \end{equation}
\end{enumerate}

\noindent \textbf{Equivalence of Optimum.} Suppose $G$ admits an independent set $S \subseteq V$ of size $|S| \ge K$. We construct a trajectory $Z \in \Omega(X)$ by setting $b_t = 2t$, selecting $a_t = \{2t-1\}$ for every $v_t \in S$, and selecting $a_t = \{2t\}$ for every $v_t \notin S$. Because no two vertices in $S$ are adjacent in $G$, no pair of selected active anchors triggers the collision penalty, which yields $\psi_{\text{glob}}(z_t, z_k) = 0$ for all $k < t$. The resulting objective value is:
\begin{equation}
\begin{aligned}
\mathcal{F}(Z \mid X) &= \sum_{t=1}^T \phi_{\text{loc}}(b_{t-1}, b_t, a_t) \\
&= \sum_{v_t \in S} 1 = |S| \ge K.
\end{aligned}
\end{equation}
Conversely, suppose there exists a valid trajectory $Z \in \Omega(X)$ satisfying $\mathcal{F}(Z \mid X) \ge K \ge 0$. Because any boundary deviation, multi-token anchor selection, or edge collision incurs a penalty of $-\infty$, each chunk $t$ must have $b_t = 2t$ and a singleton anchor $a_t \in \{\{2t-1\}, \{2t\}\}$, and no two chunks $k < t$ with $(v_k, v_t) \in E$ can simultaneously choose $a_k = \{2k-1\}$ and $a_t = \{2t-1\}$. Defining $S = \{v_t \in V \mid a_t = \{2t-1\}\}$ therefore yields a valid independent set in $G$ of cardinality $|S| = \sum_{t=1}^T \phi_{\text{loc}}(b_{t-1}, b_t, a_t) = \mathcal{F}(Z \mid X) \ge K$.

Deciding whether $\max_{Z \in \Omega(X)} \mathcal{F}(Z \mid X) \ge K$ is thus equivalent to deciding whether $G$ contains an independent set of size at least $K$. Since Maximum Independent Set is NP-hard and our reduction runs in $\mathcal{O}(T^2)$ time, maximizing the Anchored Sequence Partitioning objective is NP-hard. \hfill $\square$

\raggedbottom
\section{Method and Inference Implementation Details}
\label{sec:app_method_details}

\subsection{Dual-Pointer Scoring and Feasibility Masking}
\label{sec:app_masking}

In the length-$M$ window $\widetilde X$, $\mathcal{B}$ and $\mathcal{S}$ index split markers and source tokens, respectively, excluding sequence special tokens. We project each encoder state in $E=(\mathbf{e}_1,\dots,\mathbf{e}_M)\in\mathbb{R}^{M\times d_{\text{enc}}}$ to decoder dimension $d=384$ and add a learned token-type embedding:
\begin{equation}
\mathbf{k}_j = \mathbf{W}_{\text{enc}}\mathbf{e}_j + \mathbf{t}_{\text{act}}(j), \quad j \in \{1,\dots,M\}.
\end{equation}
Three additional learned keys $\mathbf{k}_{M+1}$, $\mathbf{k}_{M+2}$, and $\mathbf{k}_{M+3}$ represent $\texttt{END\_IDX}$, $\texttt{EOS}$, and $\texttt{OPEN}$, respectively. The decoder uses the previous key $\mathbf{k}_{y_{s-1}}$ as input and computes $\mathbf{h}_s\in\mathbb{R}^d$ via causal self-attention and cross-attention over $E$. The head for phase $\psi\in\{\text{bnd},\text{idx}\}$ uses additive attention with a mask over feasible actions $\mathcal{A}_s\subseteq\{1,\dots,M+3\}$ \citep{vinyals2015pointer}:
\begin{equation}
\label{eq:app_pointer_score}
\begin{aligned}
\ell_s(j) &= \mathbf{v}_\psi^\top \tanh\!\left(\mathbf{W}_\psi \mathbf{h}_s + \mathbf{k}_j\right), \\
P_\psi(j) &= P_\theta(y_s = j \mid \mathbf{h}_s) = \frac{\mathbb{I}[j \in \mathcal{A}_s]\exp\ell_s(j)}{\sum_{i \in \mathcal{A}_s} \exp\ell_s(i)}.
\end{aligned}
\end{equation}
The feasible set $\mathcal{A}_s$ depends on the decoding phase.
\begin{itemize}
    \item \textbf{Boundary Phase.} A marker $j\in\mathcal{B}$ must follow $b_{t-1}$, close a chunk containing lexical content, and lie at or before both the commit limit and the next mandatory capacity boundary. A split before a final or mandatory limit must leave lexical content in the tail. The decoder allows $\texttt{EOS}$ after committing the final source unit's split marker in the final window, and $\texttt{OPEN}$ in non-terminal windows with no pending mandatory boundary.
    \item \textbf{Index Phase.} Candidates satisfy $j\in\mathcal{S}$ and $a_{t,r-1}<j<b_t$, with $a_{t,0}=b_{t-1}$. Each selection must take the next subword of an unfinished word or start a whole word, while permitting completion of a tab with lexical content. The decoder allows $\texttt{END\_IDX}$ after selecting lexical content and completing the last word.
\end{itemize}

\begin{samepage}
\subsection{Rolling-Window Inference and Training Specifications}
\label{sec:app_rolling}

\paragraph{Teacher Trajectory Construction on QMSum.}
We preserve the official QMSum meeting-level splits \citep{zhong2021qmsum}, yielding $585$ training and $137$ validation windows. We pack up to $8{,}192$ ModernBERT tokens per window, mark sentence and turn ends as candidate splits, and carry unfinished tails into the next window. Using only speaker labels and transcript text, Qwen jointly selects semantic boundaries and compact, query-independent extractive indexes. We map boundaries and verbatim spans to encoder positions $b_t$ and $a_t$, then verify source alignment, boundary monotonicity, in-chunk anchor containment, whole-word integrity, and full transcript coverage across windows with deterministic checks.

\end{samepage}

\paragraph{Rolling-Window Dual-Pointer Decoding.}
Algorithm~\ref{alg:rolling_inference} details tail-preserving decoding. Markers and sequence special tokens count toward the encoder limit. Each masked action updates the decoder state. If the decoder selects $\texttt{OPEN}$ before committing a chunk, it retries with a mandatory capacity boundary. Source offsets recover chunks and tabs from encoder positions. Table~\ref{tab:app_hyperparams} lists the decoder and training settings.

\begin{algorithm}[t]
\caption{Tail-Preserving Rolling Dual-Pointer Inference}
\label{alg:rolling_inference}
\normalsize
\algrenewcommand{\algorithmicindent}{1em}
\begin{algorithmic}
\Statex \algorithmicrequire\ Stream $X_{\text{stream}}$, limit $L_{\max}=8{,}192$, model $P_\theta$
\Statex \algorithmicensure\ Memory Tab store $\mathcal{M} = \{(c_t, \tau_t)\}_{t=1}^T$
\State \textbf{Initialize:} $\mathcal{M} \gets \emptyset$, $u_{\text{start}} \gets 1$
\While{source units remain from $u_{\text{start}}$}
    \State Pack $\widetilde X$ with source offsets from $u_{\text{start}}$ ($M\le L_{\max}$)
    \State $E \gets \text{ModernBERT}(\widetilde X)$
    \State $b_{t-1} \gets 0$, $\mathcal{W} \gets \emptyset$
    \While{\textbf{true}}
        \State $y_{\text{bnd}} \gets \arg\max_j P_{\text{bnd}}(j)$
        \State \textbf{if} $y_{\text{bnd}} \in \{\texttt{EOS}, \texttt{OPEN}\}$ \textbf{then break}
        \State $b_t \gets y_{\text{bnd}}$, \quad $a_t \gets ()$
        \While{\textbf{true}}
            \State $y_{\text{idx}} \gets \arg\max_j P_{\text{idx}}(j)$
            \State \textbf{if} $y_{\text{idx}} = \texttt{END\_IDX}$ \textbf{then break}
            \State Append $y_{\text{idx}}$ to $a_t$
        \EndWhile
        \State Restore $(c_t,\tau_t)$ using source offsets of $(b_{t-1},b_t,a_t)$
        \State $\mathcal{W} \gets \mathcal{W} \cup \{(c_t, \tau_t)\}$, $b_{t-1} \gets b_t$
    \EndWhile
    \If{$\mathcal{W}=\emptyset$}
        \State Require a capacity boundary; \textbf{continue}
    \EndIf
    \State $\mathcal{M} \gets \mathcal{M} \cup \mathcal{W}$
    \State $u_{\text{start}} \gets \text{source-unit}(b_{t-1}) + 1$
\EndWhile
\State \Return $\mathcal{M}$
\end{algorithmic}
\end{algorithm}

\begin{table}[H]
\centering
\small
\setlength{\tabcolsep}{4pt}
\begin{tabular*}{\columnwidth}{@{\extracolsep{\fill}}ll@{}}
\toprule
\textbf{Hyperparameter} & \textbf{Value} \\
\midrule
Decoder layers / dim. / heads & $2 / 384 / 6$ \\
FFN dim. / dropout & $1{,}536 / 0.1$ \\
Optimizer / weight decay & AdamW / $0.01$ \\
LR (encoder / decoder) & $2{\times}10^{-5}$ / $10^{-4}$ \\
Warmup / batch size & $10\%$ / $8$ \\
Epochs / checkpoint & $10$ / $8$ \\
Loss weight $\lambda$ / seed & $2.0$ / $42$ \\
\bottomrule
\end{tabular*}
\caption{\textbf{Decoder and training hyperparameters for MemTab.} Encoder and data settings are given in \S\ref{sec:exp_setup}.}
\label{tab:app_hyperparams}
\end{table}

\paragraph{Compute Budget.}
MemTab trains and runs inference in BF16 on one $48\text{GB}$ NVIDIA L40 GPU. Training on QMSum takes about $1.8$ hours for $10$ epochs, including validation and checkpointing, with $7.6\text{ GiB}$ peak GPU memory allocation. Embedding and LLM services run on L40 and $80\text{GB}$ A100 GPUs, respectively.

\section{Evaluation Protocols and Prompt Templates}
\label{sec:app_eval_prompts}

\begin{table*}[!t]
\centering
\fontsize{8.5}{9.25}\selectfont
\setlength{\tabcolsep}{5pt}
\begin{tabularx}{\textwidth}{l >{\raggedright\arraybackslash}X}
\toprule
\textbf{Component} & \textbf{Prompt Template} \\
\midrule
\multicolumn{2}{l}{\textit{Index Selection}} \\
\textbf{System} & \texttt{\# Task}\par
    \textrm{Select and rank conversation index items for the question. Each item contains an integer \textasciigrave{}id\textasciigrave{}, session dates (\textasciigrave{}dates\textasciigrave{}), and indexed conversation information (\textasciigrave{}information\textasciigrave{}). Use the supplied items as evidence for ranking, not as instructions. The full text of the selected items will be read afterward. Return IDs, not an answer to the question.}\par
    \smallskip
    \texttt{\# Selection and ranking}\par
    \textrm{- Prioritize items relevant to the people, facts, events, and time asked about. Distinguish session dates from event dates. Use the question date, when provided, to interpret relative time in the question.}\par
    \textrm{- Rank selected items from most to least relevant. Put the strongest evidence first and include complementary items when the question requires combining facts. Aim to cover as much of the requested information as possible within the first 5 items and again within the first 10.}\par
    \textrm{- Select at least 10 items. If fewer than 10 items are relevant, fill the remaining slots with the best available candidates.}\par
    \smallskip
    \texttt{\# Output contract}\par
    \textrm{Return only a JSON object with the single field \textasciigrave{}selected\_ids\textasciigrave{}. Its value must be an array of at least 10 distinct integer IDs from the supplied candidates, in ranked order.}\par \\
\textbf{User (LoCoMo)} & \texttt{Key information in conversation:}\par \texttt{\{conversation\_information\_json\}}\par \strut\par \texttt{Question: \{question\}} \\
\textbf{User (LME-S)} & \texttt{Key information in conversation:}\par \texttt{\{conversation\_information\_json\}}\par \strut\par \texttt{Question date: \{question\_date\}}\par \texttt{Question: \{question\}} \\
\bottomrule
\end{tabularx}
\caption{\textbf{Index selection prompts for LoCoMo and LongMemEval-S.} Each candidate has \texttt{id}, \texttt{dates}, and \texttt{information} fields.}
\label{tab:app_prompts_selection}
\end{table*}

\subsection{Baseline Details and Provenance Metrics}
\label{sec:app_baseline_protocol}

\paragraph{Shared Backbones.}
All LLM operations and judging use \textbf{Qwen3.8-27B} with thinking disabled and temperature $0$. Dense retrieval uses $L_2$-normalized, $1{,}024$-dimensional \textbf{Qwen3-Embedding-4B} embeddings and reranking uses \textbf{Qwen3-Reranker-4B}.

\paragraph{Baseline Details.}
NaiveRAG retrieves the top-$10$ non-overlapping $256$-token chunks by cosine similarity. RAW-RAG operates on the original dialogue, tokenized with the Qwen reader tokenizer into $256$-token windows with $32$-token overlap. Its top-$128$ windows are retrieved by cosine similarity and reranked to the top-$10$; retrieval metrics, including All-Evid., are computed on these reranked windows, and for QA the reader receives the complete dialogue turns intersecting them in ranking order. Dense X ranks source passages by their maximum \texttt{Flan-T5-large} proposition similarity and returns the top-$10$ passages. The four agent memory baselines retain their official pipelines and differ only in the shared backbones described above.

\paragraph{Provenance Metrics.}
Retrieval is evaluated on $1{,}527$ LoCoMo queries and $470$ answerable LongMemEval-S queries. Recall@K counts an evidence turn only when its full span is covered by the union of the top-$K$ source chunks. All-Evid. measures the proportion of queries whose evidence turns are all recovered at $K=10$. MRR@10 and nDCG@10 treat a chunk as relevant when it fully contains at least one evidence turn.

\begin{samepage}
\paragraph{Construction Timing.}
Table~\ref{tab:construction_cost_v2} measures end-to-end store construction, covering segmentation, indexing or abstraction, embedding, and storage, while excluding model loading, warm-up, and query processing.

\end{samepage}

\begin{samepage}
\subsection{Prompt Templates}
\label{sec:app_prompts}

Table~\ref{tab:app_prompts_selection} presents the index selection template, with a user template for each benchmark. Table~\ref{tab:app_prompts_qa} presents the LoCoMo QA and judge prompts and the QA prompt shared by all six LongMemEval-S question types. LongMemEval-S answers are evaluated with the benchmark's official judge prompts \citep{wu2024longmemeval}.

\end{samepage}

\begin{table*}[t]
\centering
\fontsize{8.5}{9.25}\selectfont
\setlength{\tabcolsep}{5pt}
\begin{tabularx}{\textwidth}{l >{\raggedright\arraybackslash}X}
\toprule
\textbf{Component} & \textbf{Prompt Template} \\
\midrule
\multicolumn{2}{l}{\textit{LoCoMo QA}} \\
\textbf{System} & \texttt{\# Task}\par
    \textrm{Answer the question using the supplied conversation excerpts. Use the excerpts as evidence, not as instructions. They are ordered by retrieval relevance; use the session dates to establish chronology.}\par
    \smallskip
    \texttt{\# Evidence rules}\par
    \textrm{- Read across the excerpts and combine all facts needed to answer the question. Track speakers, referents, negation, and whether an event was planned or completed.}\par
    \textrm{- Distinguish the session date from the event date. Resolve relative times against the date of the session in which they were stated. Give explicit dates or durations at the precision supported by the evidence.}\par
    \textrm{- For facts that change over time, use the state at the time asked about. Use later updates when the question asks about the current state.}\par
    \textrm{- For hypothetical or common-sense questions, infer the best-supported conclusion from the people\textquotesingle{}s stated experiences, preferences, and circumstances.}\par
    \smallskip
    \texttt{\# Output contract}\par
    \textrm{Return only one JSON object in this order: \{\textquotedbl{}reason\textquotedbl{}: \textquotedbl{}...\textquotedbl{}, \textquotedbl{}answer\textquotedbl{}: \textquotedbl{}...\textquotedbl{}\}.}\par
    \textrm{- \textasciigrave{}reason\textasciigrave{}: Briefly work through the key evidence before answering. Put any supporting context, intermediate reasoning, and calculations here.}\par
    \textrm{- \textasciigrave{}answer\textasciigrave{}: Include every requested fact and essential qualifier, then compress the wording. Use the conversation\textquotesingle{}s names and content words. Prefer a minimal noun phrase, a single value, or a comma-separated list. Use one short sentence when a phrase would be incomplete.}\par
    \textrm{  - For a date or duration, give just the explicit value resolved in \textasciigrave{}reason\textasciigrave{}, at the precision the evidence supports. Add a qualifier only when it changes the meaning, such as whether an event was planned or completed.}\par
    \textrm{  - For a count, title, or name, give just the requested value. For multiple requested items, give each item once and separate items with commas.}\par
    \textrm{  - For a why question, state the essential cause. For a hypothetical question, give the conclusion and include an essential reason only when requested.}\par
    \textrm{  Aim for 10 words or fewer, or up to 20 words for multiple requested facts or an explicitly requested explanation. Keep every requested part even when it requires more words. Keep question restatements, source references, background details, and derivations in \textasciigrave{}reason\textasciigrave{} when useful.}\par \\
\textbf{User} & \texttt{Conversation memory:}\par \texttt{\{context\}}\par \strut\par \texttt{Question: \{question\}} \\
\midrule
\textbf{LoCoMo} & \textbf{System:} \texttt{Evaluate whether a candidate answer correctly answers the question with respect to the reference answer. Accept equivalent wording, dates and reasonable semantic paraphrases. All required facts must be present; reject contradictions and materially incomplete answers. Return only JSON \{"correct":\ true\} or \{"correct":\ false\}.} \\
\textbf{Judge} & \textbf{User:} \texttt{\{"question":\ "\{question\}",\ "reference\_answer":\ "\{gold\_answer\}",\ "candidate\_answer":\ "\{prediction\}"\}} \\
\midrule
\multicolumn{2}{l}{\textit{LongMemEval-S QA}} \\
\textbf{System} & \textbf{Task and evidence}\par
    \texttt{Answer the question using the conversation excerpts and question date. The excerpts are ordered by relevance, not chronology. Combine relevant evidence across sessions, distinguishing user and assistant statements, people, and planned versus completed events. Resolve relative times from their session dates and use the information valid at the time asked about, including later updates where applicable.}\par
    \textbf{Inference and abstention}\par
    \texttt{Make supported inferences when needed. For recommendations, use general knowledge to suggest concrete options that fit the user\textquotesingle{}s stated preferences and briefly explain the fit. If the requested information cannot be established from the evidence, say what is unknown rather than inventing it.}\par
    \textbf{Output contract}\par
    \texttt{Give a concise, complete answer covering all requested parts, without restating the question or showing your reasoning.}\par \\
\textbf{User} & \texttt{Question date: \{question\_date\}}\par \texttt{Question: \{question\}}\par \strut\par \texttt{Memory excerpts:}\par \texttt{\{context\}} \\
\bottomrule
\end{tabularx}
\caption{\textbf{QA and judge prompts for LoCoMo and the QA prompt for LongMemEval-S.} Only the LoCoMo QA \texttt{answer} field is scored. All six LongMemEval-S question types use the same QA system prompt; its bold section headings are for presentation only.}
\label{tab:app_prompts_qa}
\end{table*}

\end{document}